\pdfoutput=1

\documentclass[11pt]{article}

\usepackage[]{EMNLP2022}

\usepackage{times}
\usepackage{latexsym}

\usepackage[T1]{fontenc}

\usepackage[utf8]{inputenc}

\usepackage{microtype}

\usepackage{inconsolata}

\usepackage{graphicx}

\title{SAIDS: A Novel Approach for \\ Sentiment Analysis Informed of Dialect and Sarcasm}

\author{Abdelrahman Kaseb \and Mona Farouk \\
        Computer Engineering, Cairo University \\ Giza, Egypt \\
        \texttt{\{abdelrahman.kaseb,mona\_farouk\}@eng.cu.edu.eg}}

\begin{document}
\maketitle
\begin{abstract}
Sentiment analysis becomes an essential part of every social network, as it enables decision-makers to know more about users' opinions in almost all life aspects. 
Despite its importance, there are multiple issues it encounters like the sentiment of the sarcastic text which is one of the main challenges of sentiment analysis.
This paper tackles this challenge by introducing a novel system (SAIDS) that predicts the sentiment, sarcasm and dialect of Arabic tweets. SAIDS uses its prediction of sarcasm and dialect as known information to predict the sentiment. 
It uses MARBERT as a language model to generate sentence embedding, then passes it to the sarcasm and dialect models, and then the outputs of the three models are concatenated and passed to the sentiment analysis model.
Multiple system design setups were experimented with and reported. SAIDS was applied to the ArSarcasm-v2 dataset where it outperforms the state-of-the-art model for the sentiment analysis task. 
By training all tasks together, SAIDS achieves results of 75.98 FPN, 59.09 F1-score and 71.13 F1-score for sentiment analysis, sarcasm detection, and dialect identification respectively.
The system design can be used to enhance the performance of any task which is dependent on other tasks.  
\end{abstract}

\section{Introduction}
Sentiment analysis (SA) is one of the main tasks in the natural language processing (NLP) field. It is used for opinion mining which supports decision-makers. 
Working on sentiment analysis starts relatively early, for example, \citet{pang-etal-2002-thumbs} analysed the sentiment to positive and negative in movie reviews.
Following this paper, sentiment analysis becomes one of the most important topics in NLP, especially with the increasing number of reviews on websites and social media platforms. 
Since then, a lot of work has been done in English sentiment analysis, while Arabic has relatively much less.
Since \citet{1361684.1361685} started their work on Arabic SA, multiple researchers also began theirs. Now there are well-known Arabic SA models like \citep{978-3-319-99740-7_12, 6716448, ABUFARHA2021102438, ELSHAKANKERY2019163}. 
Of course, working with Arabic has many challenges, one of the most challenging issues is the complex morphology of the Arabic language \citep{Kaseb2016ArabicSA, ABDULMAGEED2019291}. 
Another challenge is the variety of Arabic dialects \citep{ABDULMAGEED2019291}. 
Moreover, one of the well-known challenges in SA for all languages is sarcasm, as the sarcastic person uses words and means the opposite of it. 
For example, "I'd really truly love going out in this weather!", does it reflect a positive or negative sentiment? because of the sarcasm, we cannot judge the sentiment correctly.  \par

Several related works tackle English sarcasm detection with sentiment analysis \citep{oprea-magdy-2020-isarcasm, abercrombie-hovy-2016-putting, barbieri-etal-2014-modelling}. 
On the other hand, there are only a few works on both sentiment and sarcasm in Arabic. There are two shared tasks on sarcasm detection \citep{inproceedings_ghanem_2019}, but for both sarcasm and sentiment there was only one shared task \citet{abu-farha-etal-2021-overview} but each sub-task is independent, meaning that participating teams can submit a different model for each task. 
Some participants used the same model for both sentiment and sarcasm \citep{el-mahdaouy-etal-2021-deep}. \par

Instead of training sentiment independently of sarcasm, this work introduces a new model architecture that works with multi-task training which trains both at the same time. 
There are other additions to the proposed architecture; firstly, it trains with dialect also. Secondly, the sarcasm and dialect that are initially predicted are used in the prediction of the sentiment. 
In other words, the sentiment model is informed by the sarcasm and dialect model output. The contributions offered by this work are:

\begin{itemize}
\item Design a novel model architecture that can be used for a complicated task that is dependent on another task, e.g. sentiment analysis which is dependent on sarcasm detection.
\item Investigate the design setups for the new architecture and find the best setup that could be used.
\item Train the model on ArSarcam-v2 dataset and achieve the state-of-the-art results recorded as 75.98 FPN on sentiment analysis.
\end{itemize}

This paper is organized as follows Section \ref{sec:relatedwork} shows the related work on sentiment analysis, sarcasm detection, and dialect identification. Section \ref{sec:Dataset} describes the dataset used in this work and shows data statistics. Section \ref{sec:system} describes SAIDS model and all the design setups. Section \ref{sec:resutls} shows the experimental results and finally section \ref{sec:Conclusion} concludes the work.

\section{Related Work}
\label{sec:relatedwork}
SAIDS works on three tasks sentiment analysis, sarcasm detection, and dialect identification. In this section, the existing methods for each task are discussed. 
\subsection{Sentiment Analysis}
Arabic sentiment analysis started with \citet{1361684.1361685} work. 
Since then, it is developed by multiple researchers. 
In the beginning, the main focus was on modern standard Arabic (MSA), but over time the researchers start to focus on dialectal Arabic \citep{mourad-darwish-2013-subjectivity, Kaseb3ArabicSA}. \par 

Regarding the datasets, based on \citet{abs-2110-06744}, there are more than fifty datasets for sentiment analysis, including \citet{3471287.3471293, Kaseb2ArabicSA, kiritchenko-etal-2016-semeval, rosenthal-etal-2017-semeval, Elmadany2018ArSASA} datasets. 
Because of the massive number of datasets, there are a massive number of system approaches for Arabic sentiments \citep{abu-farha-magdy-2019-mazajak, 978-3-319-99740-7_12, el-beltagy-etal-2017-niletmrg}. Based on \citet{ABUFARHA2021102438} comparative study, using the word embedding with deep learning models outperform, the classical machine learning models and the transformer-based models outperform both of them. 
There is a reasonable number of Arabic transformer-based models like AraBERT \citep{antoun2020arabert} and MARBERT \citep{abdul-mageed-etal-2021-arbert} which are used by most Arabic sentiment analysis papers. \par

\subsection{Sarcasm Detection}
Unlike Arabic sentiment analysis, Arabic sarcasm detection has not gotten much attention yet. 
Only a few research works tackle the problem and still there is an obvious shortage of the Arabic sarcasm datasets, like \citet{KAROUI2017161, abu-farha-etal-2022-semeval}.
\citet{abbes-etal-2020-daict} collected a dataset for sarcastic tweets, they used hashtags to collect the dataset for example  \#sarcasm. 
Then, they built multiple classical machine learning models SVM, Naive Bayes, and Logistic Regression, the best F1-score was 0.73. \par

After that, \citet{inproceedings_ghanem_2019} organized a shared task in a workshop on Arabic sarcasm detection. 
They built the dataset by collecting tweets on different topics and using hashtags to set the class. 
An additional step was added, by sampling some of the datasets and manually annotating them. 
In this shared task, eighteen teams were working on sarcasm detection.
\citet{article_Khalifa_ensable} was the first team and achieved a 0.85 F1-score. \par

Then \citet{abu-farha-etal-2021-overview} made two tasks based on the ArSarcasm-v2 dataset; sentiment analysis and sarcasm detection. They have 27 teams participating in the workshop, the top teams achieved 62.25 F1-score and 74.80 FPN for sarcasm detection and sentiment analysis respectively. \par

\subsection{Dialect Identification}
Arabic dialect identification is an NLP task to identify the dialect of a written text. 
It can be on three levels, the first level is to identify MSA, classical Arabic (CA), and dialectical Arabic \citep{McWhorter}. The second level is to identify the dialect based on five main Arabic dialects EGY, LEV, NOR, Gulf, and MSA \citep{el-haj-2020-habibi, khalifa-etal-2016-large, 2632188.2632207, al-sabbagh-girju-2012-yadac, egan-2010-cross}. The third level is to identify the country-level dialect \citep{abdul-mageed-etal-2020-nadi}.  \par

Regarding the datasets, there are datasets more than twenty Arabic datasets labeled with dialect. One of the most popular datasets is  MADAR \citep{bouamor-etal-2018-madar} where the data is labeled at the city-level for 25 Arab cities.
\citet{abdul-mageed-etal-2020-nadi} built a shared task to detect the dialect, they published three different shared tasks. In the 2020 task, sixty teams participated, and the best results were 26.78 and 6.39 F1-score in the country-level and the city-level dialects respectively. \par

\section{Dataset}
\label{sec:Dataset}
ArSarcasm-v2 \citep{abu-farha-etal-2021-overview} is the main dataset used in this work, it was released on WANLP 2021 shared task for two tasks sarcasm and sentiment analysis. 
It has about 15k tweets and is divided into 12k for training and 3k for testing, the same test set, as released on WANLP 2021, was used. 
Each tweet was labelled for the sentiment (positive (POS), neutral (NEU), and negative (NEG)), sarcasm (true, and false), and dialect (MSA, Egypt (EGY), Levantine (LEV), Maghreb (NOR), and Gulf). 
The authors of the dataset annotate it using a crowd-sourcing platform. 
This dataset originally consisted of a combination of two datasets, the first one is ArSarcasm \citep{abu-farha-magdy-2020-arabic} and the second one is DAICT \citep{abbes-etal-2020-daict}, \citet{abu-farha-etal-2021-overview} merged the two datasets. \par

\subsection{Dataset Statistics}
In this subsection, we introduce some dataset statistics that motivated us to work on SAIDS.
The ArSarcasm-v2 dataset has 15,548 tweets, 3000 tweets are kept for testing and the rest of the tweets for training. 
Table \ref{tab:dataset_train_examples} shows the number of examples for all task labels on the training set, as we can see, most of the data is labeled as MSA and non-sarcastic in dialect and sarcasm respectively. \par

\begin{table}[htbp]
\centering
\begin{tabular}{ll|c}
\hline
\textbf{Task} & \textbf{Label} & \textbf{Count}\\
\hline
\hline
\textbf{Sentiment} & Positive &  2,180 \\
& Neutral & 5,747 \\
& Negative & 4,621 \\
\hline
\hline
\textbf{Sarcasm} & Sarcastic &  2,168 \\
& Non-sarcastic & 10,380 \\
\hline
\hline
\textbf{Dialect} & MSA &  8,562 \\
& EGY & 2,675 \\
& Gulf & 644 \\
& LEV & 624 \\
& NOR & 43 \\
\hline
\hline
\textbf{Total} &  & 12,548 \\
\hline
\end{tabular}
\caption{Number of labels of sentiment, sarcasm and dialect on the training set}
\label{tab:dataset_train_examples}
\end{table}

The relationship between sentiment labels and both sarcasm and dialect independently can be shown from Table  \ref{tab:sarcasm_sentiment_dialect}. 
For the sentiment/sarcasm part, we can see that about 90 percent of sarcastic tweets are sentimentally labeled as negative, and about 50 percent of non-sarcastic tweets are sentimentally labeled as neutral. 
On the other hand, for the sentiment/dialect part, we can see that about 50 percent of MSA tweets are sentimentally labeled as neutral and about 50 percent of EGY tweets are sentimentally labeled as negative. 
From this table, we can conclude that the information we can get on sarcasm and dialect will benefit the sentiment analysis task. \par

\begin{table}[htbp]
\centering
\begin{tabular}{l|ccc}
\hline
& \textbf{POS} & \textbf{NEU} & \textbf{NEG}\\
\hline
\hline
\textbf{Non-sarcastic} & 2,122 & 5,576 & 2,682 \\ 
\textbf{Sarcastic}  & 58 & 171 & 1,939\\
\hline
\hline
\textbf{MSA} & 1,405  & 4,486  & 2,671 \\ 
\textbf{EGY} & 506 & 793 & 1,376 \\ 
\textbf{Gulf} & 121  & 259 & 264 \\ 
\textbf{LEV} & 142 & 197 & 285 \\ 
\textbf{NOR} & 6 & 12  & 25 \\ 
\hline

\end{tabular}
\caption{Cross tabulation between sentiment labels and both sarcasm and dialect labels on the training set}
\label{tab:sarcasm_sentiment_dialect}
\end{table}

Table \ref{tab:sarcasm_dialect} shows the percentage of sarcastic tweets on each dialect. 
As the number of NOR tweets is limited, its percentage is not reliable, so we can see that Egyptians' tweets are the most sarcastic.
This supports the facts from table \ref{tab:sarcasm_sentiment_dialect} that most EGY tweets are negative and most of the sarcastic tweets are negative tweets. \par

\begin{table}[htbp]
\centering
\begin{tabular}{l|c}
\hline
\textbf{Dialect} & \textbf{Sarcasm percentage}\\
\hline
\hline
\textbf{MSA} & 10.83 \% \\ 
\textbf{EGY} & 34.77 \% \\ 
\textbf{Gulf} & 24.38 \% \\ 
\textbf{LEV} & 22.12 \% \\ 
\textbf{NOR} & 34.88 \% \\ 
\hline
\end{tabular}
\caption{Percentage of sarcastic tweets for each dialect on the training set}
\label{tab:sarcasm_dialect}
\end{table}

\section{Proposed System}
\label{sec:system}
This section presents a detailed description of the proposed system. 
SAIDS learns sentiment analysis, sarcasm detection, and dialect identification at the same time (multi-task training), in addition, it uses the sarcasm detection and dialect outputs as an additional input to the sentiment analysis model which is called "informed decision". 
SAIDS decides the sentiment class using the information of sarcasm and dialect class which are both outputs itself. 
The main idea behind SAIDS is based on analyzing the dataset statistics, as shown in section \ref{sec:Dataset}, which says that most sarcastic tweets are classified as negative tweets and most MSA tweets are classified as neutral tweets. \par 

\subsection{System Architecture}
\begin{figure}[htbp]
\centerline{\includegraphics[height=5cm]{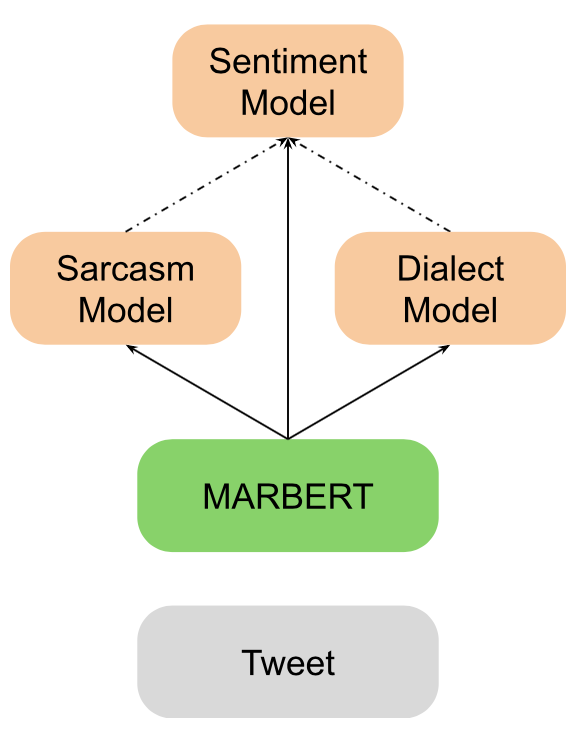}}
\caption{SAIDS architecture}
\label{system_arhc}
\end{figure}

Figure \ref{system_arhc} shows the SAIDS architecture. 
The architecture consists of four main modules, the first module is MARBERTv2 \citep{abdul-mageed-etal-2021-arbert}, it is a transformer-based model, its input is the tweet, and its output is a sentence embedding which is a vector of length 768. 
The second module is the "Sarcasm Model", it is a binary classifier for sarcasm, its input is the sentence embedding, and its output is two values one for sarcastic tweets and another for non-sarcastic tweets. 
The third module is the "Dialect Model", which is identical to the "Sarcasm Model" except that it outputs five classes (EGY, LEV, NOR, Gulf, and MSA).
The fourth module is the "Sentiment Model", it is a classifier for sentiment, its input is the concatenation of the sentence embedding, sarcasm model outputs and dialect model outputs. \par

The loss function used is Cross-Entropy for sentiment and dialect. Of course, since sarcasm is binary, we used binary Cross-Entropy for it. \par 

\subsection{Training Setups}
This subsection describes the multiple setups that were used to arrive at the best model performance. The experiments carried out utilized multiple setups regarding the architecture and the training strategies. \par

\textbf{Modules Architecture} 
Multiple architectures were tested for the "Sentiment Model", "Sarcasm Model" and "Dialect Model". 
As a proof of concept for the idea, we first built a simple random forest model in each task model (random forest version).
For the real scenario, we used multi-layer neural network (MNN) models. 
The first and the simplest is one output layer model and zero hidden layers. 
The second is one or two hidden layers, then the output layer. 
The third is one or two hidden layers the output of the module is the output of the hidden layer, which means that "Sentiment Model" inputs is not the output layer of the "Sarcasm Model" but the last hidden layer of it. 
The fourth setup is to concatenate the last hidden layer with the output layer and then pass it to "Sentiment Model".\par

\textbf{What Should Be Informed} 
The SAIDS architecture Figure \ref{system_arhc} shows that the "Sentiment Model" inputs are "Sarcasm Model" and "Dialect Model" outputs but we experimented with multiple settings in this part;
sentiment analysis informed of sarcasm only, dialect only, and both sarcasm and dialect. \par

\textbf{Limited Backpropagation} 
We limited the backpropagation over the dotted lines in Figure \ref{system_arhc}. 
It is used to ensure that the "Sarcasm Model" and the "Dialect Model" learn their main target correctly. 
When the model predicts sentiment incorrectly, its loss propagates directly to the MARBERTv2 model via the solid line and does not propagate via the dotted lines. 
Also, we evaluate SAIDS without limiting backpropagation which means the loss propagates everywhere, and with partial limiting. 
The partial limiting can be only set when the "Sarcasm Model" has hidden layers. 
We then limit the backpropagation through the sarcasm model's output layer but propagate it through the hidden layers. \par

\textbf{Activation Function} 
The experiments were carried out with Softmax as the activation function for the output of all modules.
However, for the sake of comparison, we run the training without Softmax for the modules outputs, which means that the values are not from one to zero.  \par

\textbf{Task By Task Training} 
As we train all the three tasks together with the same model, we experimented to train the first layer models, "Sarcasm Model" and "Dialect Model", for some epochs first, then train the full system together for multiple epochs. 
The motivation behind this idea is that as long as the first layer models work correctly, the sentiment analysis will correspondingly work correctly. 
We train in multiple orders like alternating between first layer models and full system and so on. \par

\textbf{Other Training Parameters} 
In our experiments, we built SAIDS and used the MARBERTv2 model provided by HuggingFace’s transformers library \citep{wolf-etal-2020-transformers}. 
Most of the experiments trained for five epochs except for a low learning rate where it was twenty epochs.
For the learning rate, we used a range from $ 1e^{-4} $ to $ 1e^{-6} $. 
The sequence was truncated to a maximum length of 128 tokens. Adam \citep{DBLPjournalscorr_KingmaB14} was used as an optimizer for all models.

\section{Results}
\label{sec:resutls}
In this section, the results achieved with SAIDS are discussed. For the sake of comparison, baselines were built for the system. 
To initially evaluate the idea itself, a random forest model baseline was built and compared with the random forest version of SAIDS. Baselines for real scenario are
baseline one (B1) which is identical to BERTModelForSequenceClassification class in HuggingFace’s \citep{wolf-etal-2020-transformers}, which takes the MARBERTv2 sentence embedding and passes it to the output layer for classification,
and baseline two (B2) which uses two hidden layers before the classification layer, the hidden layer size is equal to the "Sentiment Model" hidden layer size,
and baseline three (B3) which uses a larger hidden layer size to match the total number of trained parameters of SAIDS model.\par


For evaluation, we used the original metrics described for the dataset \citep{abu-farha-etal-2021-overview}. For sentiment analysis, the metric is the average of the F1-score for the negative and positive classes (FPN). For sarcasm detection, the metric is F1-score for the sarcastic class only (FSar). For dialect identification, we used the weighted average of the F1-score for all dialects (WFS). \par

\subsection{Results of Different Training Setups}
This subsection presents the results of the training setups and describes the best setup that was chosen for the proposed model. 
For each part of this subsection, every other setup was not changed to make the comparison fair. \par

\textbf{Modules Architecture} 
As a proof of concept for our system, the random forest (RF) model baseline was compared with the informed random forest (IRF) which is the random forest version of SAIDS. 
Table \ref{tab:randomforest} shows that IRF outperforms RF where the FPN is improved by 3 percent which is due to the proposed architecture. 
The information gained from the new inputs, "outputs of sarcasm model" and "outputs of dialect model", was 5 and 4 percent respectively. 
This means that about 10 percent of the sentiment analysis decision came from the newly added information. \par 

\begin{table}[htbp]
\centering
\begin{tabular}{l|c}
\hline
\textbf{Model} & \textbf{FPN}\\
\hline
\hline
\textbf{Random Forest} & 59.36 \\
\textbf{Informed Random Forest} & 62.34 \\
\hline
\end{tabular}
\caption{Performance comparison for the proof of concept on the validation set}
\label{tab:randomforest}
\end{table}

For the MNN architecture of the modules, multiple numbers of hidden layers were trained. At each experiment, all the modules have the same number of hidden layers.  
Table \ref{tab:hiddenlayers} shows that using zero hidden layers gives the best results. 
So no hidden layer setup was used in SAIDS. \par

\begin{table}[htbp]
\centering
\begin{tabular}{l|c}
\hline
\textbf{Model} & \textbf{FPN}\\
\hline
\hline
\textbf{0 Hidden Layer} & 75.23 \\
\textbf{1 Hidden Layer} & 74.90 \\
\textbf{2 Hidden Layer} & 74.89 \\
\hline
\end{tabular}
\caption{Performance comparison for the number of hidden layers in modules on the validation set}
\label{tab:hiddenlayers}
\end{table}

\textbf{What Should Be Informed} 
Experiments were also done to find the best features to use while analysing sentiment. 
Table \ref{tab:whatinformed} shows that using both dialect and sarcasm is better than using only one of them and of course better than not using any of them which is the baseline. 
With a quick observation, it was found out that the dialect benefits the sentiment more than the sarcasm, this can be obvious when speaking about MSA tweets because most of them are labeled as neutral on sentiment. Accordingly, sarcasm and dialect information was used in SAIDS. \par

\begin{table}[htbp]
\centering
\begin{tabular}{l|c}
\hline
\textbf{Model} & \textbf{FPN}\\
\hline
\hline
\textbf{Not Informed (B1)} & 72.40 \\
\textbf{Informed of sarcasm} & 73.67 \\
\textbf{Informed of dialect} & 74.41 \\
\textbf{Informed of sarcasm and dialect} & 75.23 \\
\hline
\end{tabular}
\caption{Performance comparison for what should be informed on the validation set}
\label{tab:whatinformed}
\end{table}

\textbf{Limited Backpropagation} 
Experiments were also done to find the best path for backpropagation to work with. "Full limit" is when the loss does not propagate through the "Sarcasm model" and "Dialect Model", "Partial limit" is when it propagates through some layers, and  "Unlimited" is when it propagates through all layers. 
The model was composed of two hidden layers while running these experiments.
Table \ref{tab:propagate} shows that "Partial limit" gets better results than the others, but on SAIDS we did not use it as we used a no hidden layer setup, so we used the "Full limit" backpropagation.

\begin{table}[htbp]
\centering
\begin{tabular}{l|c}
\hline
\textbf{Model} & \textbf{FPN}\\
\hline
\hline
\textbf{Full limit} & 74.23 \\
\textbf{Partial limit} & 74.89 \\
\textbf{Unlimited} & 72.31 \\
\hline
\end{tabular}
\caption{Performance comparison for limiting backpropagation  on the validation set}
\label{tab:propagate}
\end{table}

\textbf{Activation Function} 
For the sake of comparison, the Softmax layer was removed from the output layer of the model in the experiments. 
Table \ref{tab:activation} compares both setups, it shows that, as expected, using Softmax is better than not using it, as it quantify the probability of being sarcasm or being a certain dialect. So in SAIDS, Softmax was used on each module. \par

\begin{table}[htbp]
\centering
\begin{tabular}{l|c}
\hline
\textbf{Model} & \textbf{FPN}\\
\hline
\hline
\textbf{With Softmax} & 75.23 \\
\textbf{Without Softmax} & 72.15 \\
\hline
\end{tabular}
\caption{Performance comparison for the activation function setting on the validation set}
\label{tab:activation}
\end{table}

\textbf{Task By Task Training} 
Experiments were also done with training the three tasks together at the same time (All tasks), and multiple sets of the training sequence. 
The first is one epoch of training for sarcasm and dialect, and the rest for the full system (Seq 1).  
The second is odd epochs for sarcasm and dialect and even epochs for the full system (Seq 2). 
The third is two epochs of training for sarcasm and dialect and the rest for sentiment only (Seq 3). 
Table \ref{tab:trainseq} shows that Seq 1 performs better than the other sequences, so we used it for the final model training. \par

\begin{table}[htbp]
\centering
\begin{tabular}{l|c}
\hline
\textbf{Model} & \textbf{FPN}\\
\hline 
\hline
\textbf{All tasks} &  74.35\\
\textbf{Seq 1} &  75.23\\
\textbf{Seq 2} &  73.49\\
\textbf{Seq 3} &  73.01\\
\hline
\end{tabular}
\caption{Performance comparison for different model training sequences on the validation set}
\label{tab:trainseq}
\end{table}

\textbf{Summary of Used Setups}
SAIDS used information from sarcasm and dialect models, which are both one classification layer with no hidden layers, the sentiment loss does not propagate through sarcasm and dialect models, and the Softmax activation function was used on each model output. The used training sequence was one epoch of training for sarcasm and dialect, and the rest epochs for the full system.   \par

\subsection{Results comparison with literature}
SAIDS was trained and compared to the baselines we built and also the state-of-the-art models. 
Table \ref{tab:stateofartcompare} shows that SAIDS outperforms the existing state-of-the-art models on the sentiment analysis task. 
SAIDS's main task is sentiment analysis, the sarcasm detection and dialect identification are considered secondary outputs. 
Although the FSar score for SAIDS is considerably high, it is ranked third in the state-of-the-art models. 
On the other hand, most works that achieve state-of-the-art results are using different models for each task but in the proposed architecture, one model is used for both. 
The model also outputs the dialect, it achieves 71.13 percent on the weighted F1-score metric, but the literature has not reported the dialect performance so it is not included in the table.  \par

\begin{table}[htbp]
\centering
\begin{tabular}{l|cc}
\hline
\textbf{Model} & \textbf{FPN} & \textbf{FSar}\\
\hline 
\hline
\textbf{Baseline 1} & 71.60 & 58.41 \\
\textbf{Baseline 2} & 72.53 & 58.61\\
\textbf{Baseline 3} & 73.11 & 58.62\\
\textbf{\citet{el-mahdaouy-etal-2021-deep}} & 74.80 & 60.00\\
\textbf{\citet{song-etal-2021-deepblueai}} & 73.92 & \textbf{61.27}\\
\textbf{\citet{abdel-salam-2021-wanlp}} & 73.21 & 56.62\\
\textbf{\citet{wadhawan-2021-arabert}} & 72.55 & 58.72\\
\textbf{SAIDS} & \textbf{75.98} & 59.09 \\
\hline
\end{tabular}
\caption{Performance comparison for the state-of-the-art models and SAIDS on the test set}
\label{tab:stateofartcompare}
\end{table}

\section{Conclusion}
\label{sec:Conclusion}
Sentiment analysis is an important system that is being used extensively in decision-making, though it has different drawbacks like dealing with sarcastic sentences. 
In this work, we propose SAIDS which is a novel model architecture to tackle this problem. 
SAIDS essentially improves the sentiment analysis results while being informed of sarcasm and dialect of the sentence. 
This was achieved by training on the ArSarcasm-v2 dataset which is labeled for sentiment, sarcasm, and dialect. SAIDS's main target is to predict the sentiment of a tweet. 
It is trained to predict dialect and sarcasm, and then make use of them to predict the sentiment of the tweets.
This means that while the model is predicting the sentiment, it is informed of its sarcasm and dialect prediction. 
SAIDS achieved state-of-the-art performance on the ArSarcasm-v2 dataset for predicting the sentiment; 75.98 percent average F1-score for negative and positive sentiment. 
For sarcasm detection, SAIDS achieved a 59.09 percent F1-score for the sarcastic class, whereas for dialect identification it achieved a 71.13 percent weighted F1-score for all the dialects.
We believe that this model architecture could be used as a starting point to tackle every challenge in sentiment analysis. 
Not only sentiment analysis but also this is a general architecture that can be used in any context where the prediction of a task depends on other tasks. 
The idea behind the architecture is intuitive, train for both tasks and inform the model of the dependent task with the output of the independent task. \par


        \bibliographystyle{acl_natbib}

\end{document}